\documentclass{egpubl}

 \WsPaper         
 \electronicVersion 

\ifpdf \usepackage[pdftex]{graphicx} \pdfcompresslevel=9
\else \usepackage[dvips]{graphicx} \fi

\PrintedOrElectronic

\usepackage{t1enc,dfadobe}

\usepackage{egweblnk}
\usepackage{cite}
\usepackage{subcaption}
\usepackage{amsmath,amsfonts,amssymb}
\usepackage{bm}

\title[Automated Cinematography with Unmanned Aerial Vehicles]%
      {Automated Cinematography with Unmanned Aerial Vehicles}

\author[Q. Galvane \& J. Fleureau \& F.L. Tariolle \& P. Guillotel]
       {Q. Galvane\thanks{quentin.galvane@technicolor.com}, J. Fleureau, F.\,L. Tariolle
        and P. Guillotel\\
         Technicolor, France
       }

\begin{document}
	\teaser{
	\includegraphics[width=0.33\linewidth]{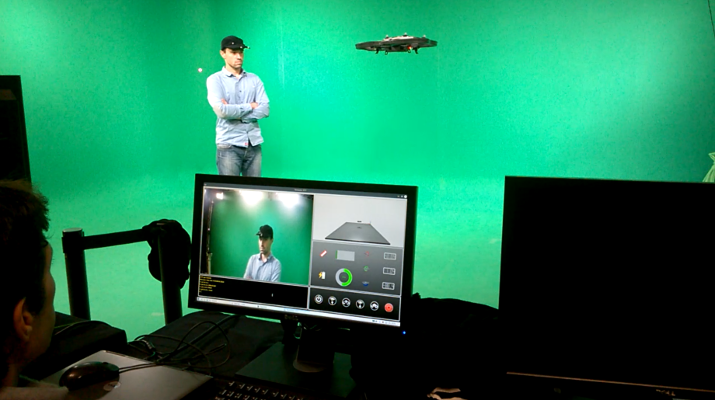}
	\includegraphics[width=0.33\linewidth]{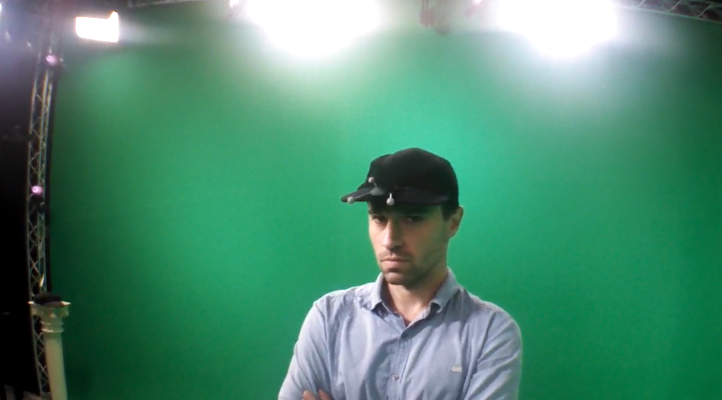}
	\includegraphics[width=0.33\linewidth]{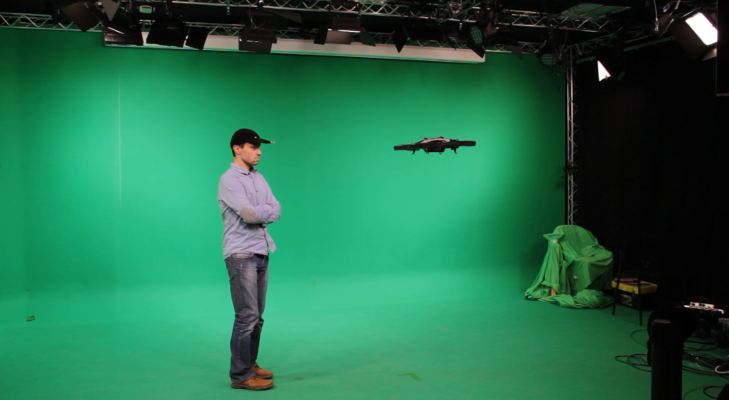}
	\centering
	\caption{Drone autonomously filming an actor given a specific cinematographic command.}
	\label{fig:teaser}
 }

\maketitle

\begin{abstract}
The rise of Unmanned Aerial Vehicles and their increasing use in the cinema industry calls for the creation of dedicated tools.
Though there is a range of techniques to automatically control drones for a variety of applications, none have considered the problem of producing cinematographic camera motion in real-time for shooting purposes.
In this paper we present our approach to UAV navigation for autonomous cinematography.
The contributions of this research are twofold:
(i) we adapt virtual camera control techniques to UAV navigation; 
(ii) we introduce a drone-independent platform for high-level user interactions that integrates cinematographic knowledge.
The results presented in this paper demonstrate the capacities of our tool to capture live movie scenes involving one or two moving actors.


\end{abstract}

\section{Introduction}


Over the last decade, the market of Unmanned Aerial Vehicles (UAV) has experienced an impressive growth. Due to their significant potential for both military and civilian use, these new vehicles -- often referred to as drones -- have drawn a lot of attention and triggered the interest of the research community.
One interesting application domain of these drones is the cinema industry. Movie-makers rely more and more on such devices to compose shots that would be impossible or otherwise extremely expensive to produce. 
When mounted with a camera and properly piloted, drones offer degrees of freedom that no other camera device could provide -- camera cranes, steadycam or camera tracks each have specific physical constraints. 
The expressiveness allowed by this novel form of camera rig however comes with a price: it requires a new set of skills and a particular expertise to pilot the drones. To produce cinematographically plausible shots, such setup usually requires two operators: a trained pilot to manually fly the drone and a camera operator to handle the framing of the shot.
Even though there has been a lot of research conducted on autonomous flight control for UAV, there is currently no literature addressing the challenge of computing cinematographic paths in real-time.

In this paper we introduce an interactive tool that allows any user to produce well composed shots by only specifying high level cinematographic commands. 
Our tool provides an intuitive interface to control drones through simple text-based interactions -- similar to orders given by a director to his cameraman -- that specify the type and the desired composition of the shot (\emph{i.e.} the placement of actors and objects on the screen). Using this information in a fully captured indoor environment (\emph{i.e.} the positions of the drones and targets are known at all times), our system performs smooth transitions between shot specifications while tracking the desired targets. 
As it no longer requires the dexterity needed to manually fly a drone nor the expertise to design camera trajectories before-hand, our tool can be used with no or very little training. It only requires basic cinematographic vocabulary. Moreover in order to allow users to train and experiment with our tool without any actor or equipment, we developed a training platform. This simulator uses the same interface and reproduces the behavior of the drones within a 3D virtual environment. 

After reviewing the related work regarding both the robotic and the cinematographic aspects of this research topic, we present an overview of our system. We illustrate the workflow and detail the tool internal processes.
In a second part we present our path planning solution, followed by the explanations on the servo control of the drone. We also give a thorough description of our framework and its functionalities.
We then detail our early experimental results.
Finally, before concluding, we present the limitations of this work and the many leads for future work and improvement.

\section{Related work}

In this section, we first give an insight on the necessary cinematographic background.
We then review the current state of the art on autonomous flight control for UAVs and associated applications. 
Finally, we address the literature related to path planning and camera control.

\subsection{Cinematography}

In the past century, based on their experience, movie-makers have defined standard practices for the cinema industry. These guidelines, first introduced in ``The 5 C's of Cinematography'' \cite{Mas1965}, define motion picture filming techniques and conventions. 
Later, many other books addressed this same issue, focusing on more specific aspects of cinematography and trying to characterize these common practices \cite{Arijon1976, Murch1986, Kat1991, Thompson1993, Thompson1998, Merc2010}.
Through this process, they have defined many stereotypical types of shots that can be described using properties such as the shot size (\emph{i.e.} Close-Up, Medium-shot, Full-shot, etc.), the profile angle (\emph{i.e.} front, right, 3/4 right, etc.), the vertical angle (\emph{i.e.} high angle, low angle or neutral) or the position of the subjects on the screen.
This \emph{grammar of the shot} was formalized by~\cite{Ronfard2013} with the Prose Storyboard Language (PSL).
The PSL is a formal language used to describe movies shot by shot, where each shot is described with a unique sentence. It covers all possible types of shots and also handles camera movements. The PSL syntax is given in Figure~\ref{fig:pslSyntax}.
This paper focusing on the placement and motion of cameras, other aspects of cinematography such as staging or lighting are not investigated here.

\subsection{Unmanned Aerial Vehicles}

There exist several categories of UAV, each with their own characteristics and capabilities: fixed-wing drones (plane-like), rotary-wing drones (helicopter-like) and flapping-wing UAV (hummingbird-like). Due to the good trade-off between the payload, control and price that they offer, rotary-wing drones are the most developed drones for the civil market. In the cinema industry especially, movie-makers exclusively use these drones as they allow to produce shots with complex camera motion and reach viewpoints inaccessible to other camera devices. Therefore, we decided to focus our research on rotary-wing drones and ignore other types of UAV, unsuited for cinema purposes.

\textbf{Autonomous target tracking.} A key element to automate the shooting process is the capability to maintain the framing on a given character and thus the capacity to track a dynamic target. 
To address this issue, \cite{Gomez2012} and \cite{Teuliere2011} devised control strategies based on computer vision. Both their solutions however heavily rely on the recognition of specific patterns. As such, these approaches would not be suitable for actors tracking.
Recently, several quadrotor manufacturers \cite{3Drsolo, Hexo, DJI} have also tackled this challenge using GPS signals and Inertial Navigation System (INS). Such tracking system however does not offer the precision needed for a satisfying control of the framing.
Finally, another common approach consists in using a motion capture system to continuously track the position and orientation of the subjects (\emph{i.e.} drones and targets) \cite{Mellinger2011}. Morevover, unlike other solutions -- mostly designed for outdoor shooting -- this approach provides the precision needed to work in an indoor environment.

\textbf{UAV navigation.}
The capacity to autonomously maneuver drones to execute a given trajectory or objective in a constrained environment is obviously essential for shooting purposes. 
Part of the research addressing this challenge focused on aggressive maneuvers in highly constrained environments \cite{Mellinger2011, Mellinger2014,Richter2013}. While impressive in terms of precision, these solutions do not provide the required stability and would produce poor camera shots.
As a ground control station, the APM Mission Planner \cite{MissionPlanner} allows users to design their own trajectories by defining waypoints on a 2D map.
With Pixhawk and the QGroundControl system, Meier \emph{et al.} \cite{Meier2011, Meier2012} go further and offer the possibility to define the trajectory in a 3D environment. While suited for camera trajectories, these approaches still require to specify the path of the drone manually before the flight.

\textbf{Cinematography with autonomous drones.}
There is currently very little literature on autonomous drones applied to cinematography.
Recently, \cite{Srikanth2014} proposed an interesting approach to control quadrotors for lighting purposes. They present a solution to automatically achieve a specific lighting effect on dynamic subjects using a drone equipped with a fixed portable light source. Their solution processes the images from a static camera to compute the 3D motion commands for the UAV.
Closer to our work, in~\cite{Joubert2015}, the authors address the challenge of autonomously performing camera shots with quadrotors. They present an interactive tool that allows users to design physically plausible trajectories by visually specifying shots. They use a virtual environment to compose and preview the shots. Their tool however remains limited to outdoor environment. It also requires to manually craft the path beforehand and does not allow to track targets in real-time.

\subsection{Path finding and automatic camera control}

Path planning has been challenging the research community for decades and the amount of literature on the matter is significant. In the robotic field especially, a number of approaches have addressed the problem. However, a large amount of this research was dedicated to ground vehicles and therefore did not fully exploit the capacities of UAV. 
Looking at the research conducted on path planning by the computer graphics community, the specific task of virtual camera control happens to be strongly related to our research topic due to the similar properties of drones and virtual cameras.
In \cite{Christie2008}, Christie \emph{et al.} review a large spectrum of the literature on intelligent camera control, mainly consisting of optimization-based or constraint-based approaches. More recently, in \cite{Lino2012}, Lino \emph{et al.} proposed an algebraic solution to the problem of placing a camera given a set of visual properties. This seminal work on static camera placement was later used to propose camera path planning solutions. In \cite {Lino2015} and \cite{Galvane2015}, the authors detail offline solutions to the problem. Closer to our problem, \cite{Galvane2013,Galvane2014} detailed a reactive approach based on steering behaviors.

\section{System overview}

\begin{figure}[ht]
	\centering
	\includegraphics[width=\linewidth]{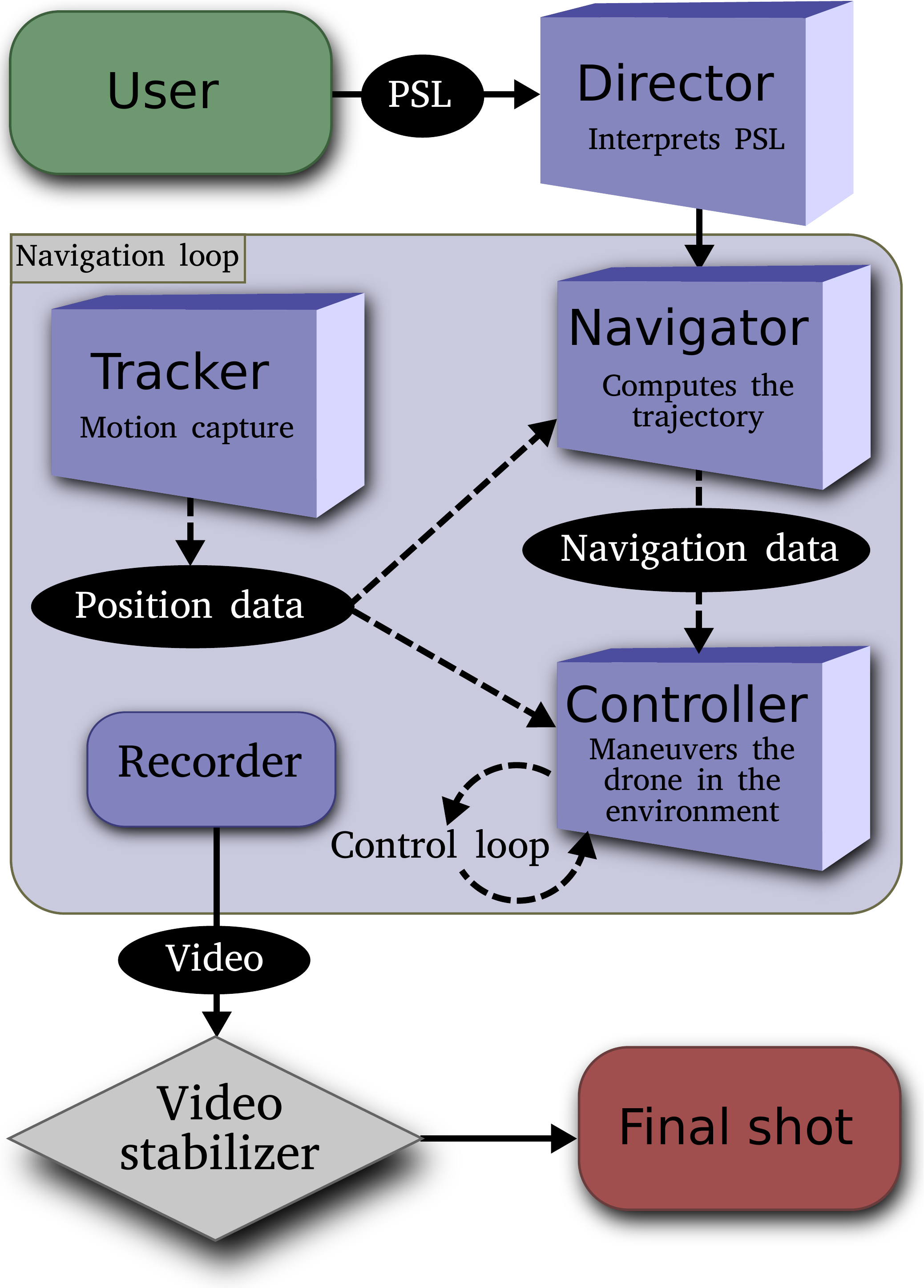}
	\caption{Overview of the system that autonomously generates a camera shot from a unique user input. The \emph{Director} interprets the command and the \emph{Navigator} handles the navigation with the \emph{Tracker} and \emph{Controller}. The recorded video is finally stabilized.}
	\label{fig:workflow}
\end{figure}

In this section, we give an overview of the system used to automatically produce camera shots through simple user interactions.
Figure~\ref{fig:workflow} details the workflow of our solution.

The first step of the process is triggered by user interactions.
Due to its straightforward grammar, we decided to use PSL sentences as the main input of our method. It allows users to easily communicate orders to the system. 
PSL commands are interpreted by a virtual \emph{Director} that extracts camera specifications and assigns them to the drones (see section~\ref{cameraSpecification}).

Then, as shown in Figure~\ref{fig:workflow}, in order to produce a cinematographic trajectory our navigation system relies on three components: a \emph{Tracker}, a \emph{Navigator} and a \emph{Controller}. 
The tracker uses a motion capture system to continuously keep track of the position and orientation of each subject in the environment (\emph{i.e.} actors and drones). 
Based on the \emph{position data} sent by the \emph{Tracker} and the camera specification given by the \emph{Director}, the \emph{Navigator} computes an initial path (see section~\ref{pathGeneration}). It then constantly send new \emph{navigation data} to guide the drone (see section~\ref{pathFollowing}). This \emph{navigation data} is used by the controller that handles the low-level control of the drones (see section~\ref{servoControl}).

Finally, once the trajectory is completed and the shot recorded, we use a video stabilizer\footnote{Adobe Premiere motion warp stabilizer.} to remove the noise induced by the drone's small deviations\footnote{This post-process is not needed when using drones equipped with camera gimbals that stabilize the camera during the flight.}.

\section{Autonomous path planning}

The input of this process is the PSL sentence given by the user ; it describes the desired framing of the shot. Our system first translates this framing into camera coordinates and then computes a feasible trajectory towards this objective.

\subsection{From PSL to shot specifications} 
\label{cameraSpecification}

\begin{figure}[ht]
	\centering
	\includegraphics[width=\linewidth]{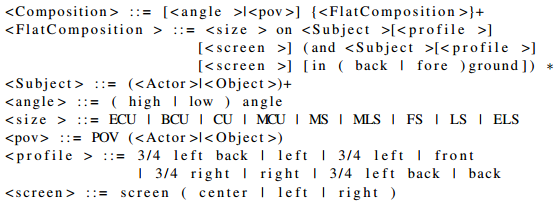}
	\caption{PSL grammar}
	\label{fig:pslSyntax}
\end{figure}

Given a set of framing properties, different optimization techniques can be used to compute actual camera configurations (\emph{i.e.} camera placement relative to the targets). 
We here rely on~\cite{Galvane2014}, based on the seminal work of~\cite{Lino2012} which gives an algebraic implementation of the problem. Camera configurations are expressed with a 2D-parametric representation, using one out of two types of manifold surfaces: a spherical surface (for single-character situations) or a toric-shaped surface (for two-character situations).
Here, the user input is a PSL shot description (which syntax is shown in Figure~\ref{fig:pslSyntax}) that represent a set of visual constraints to be satisfied. 
The spherical and toric surfaces are defined respectively by the shot size and the on-screen position of the targets. 
The optimal camera placement corresponds to the point on this surface that best satisfies the constraints.

\begin{figure}[ht]
    \centering
    \begin{subfigure}[b]{0.34\linewidth}
        \includegraphics[width=\linewidth]{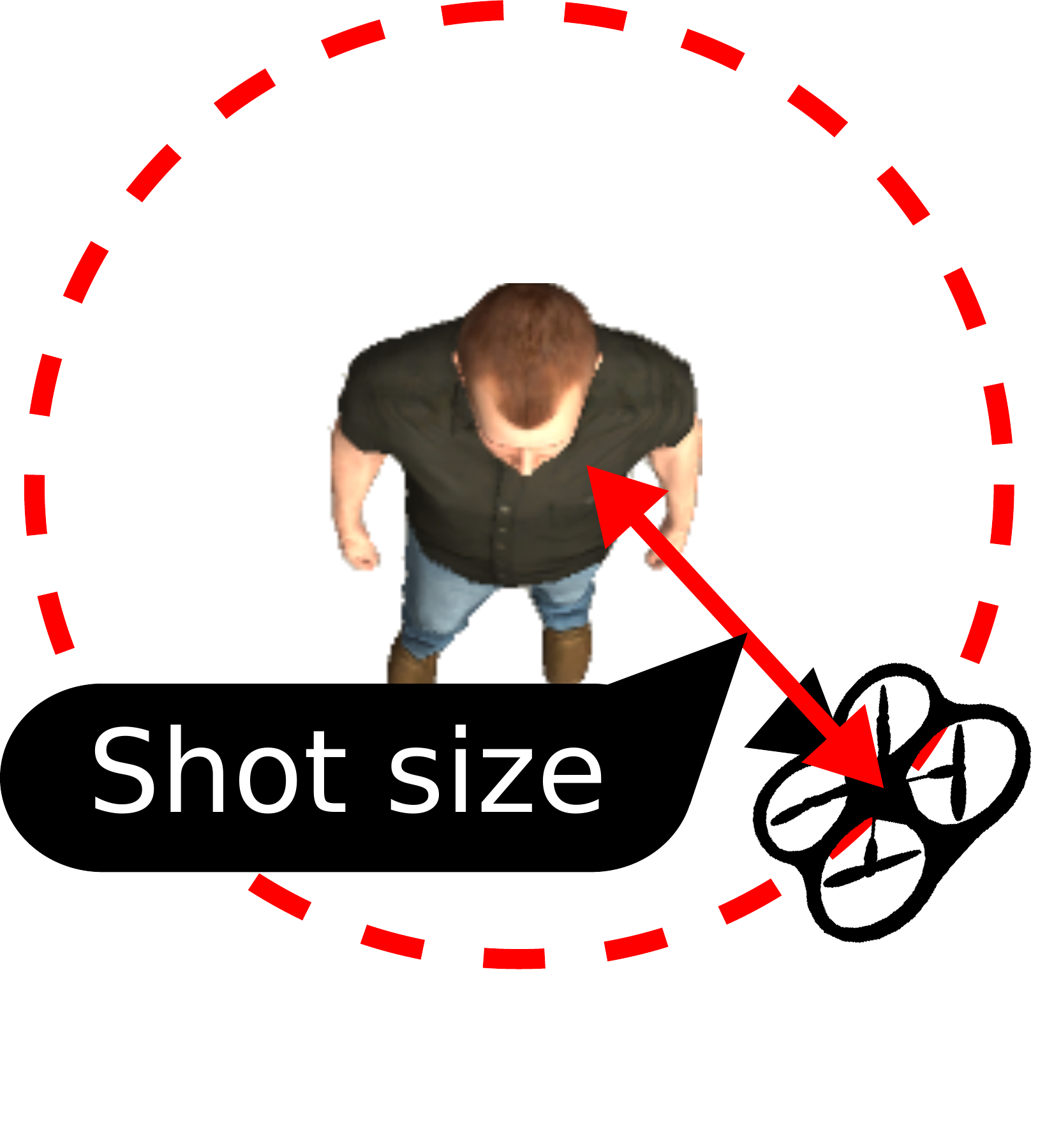}
    	\caption{}
    \end{subfigure}\hfill
    \begin{subfigure}[b]{0.64\linewidth}
        \includegraphics[width=\linewidth]{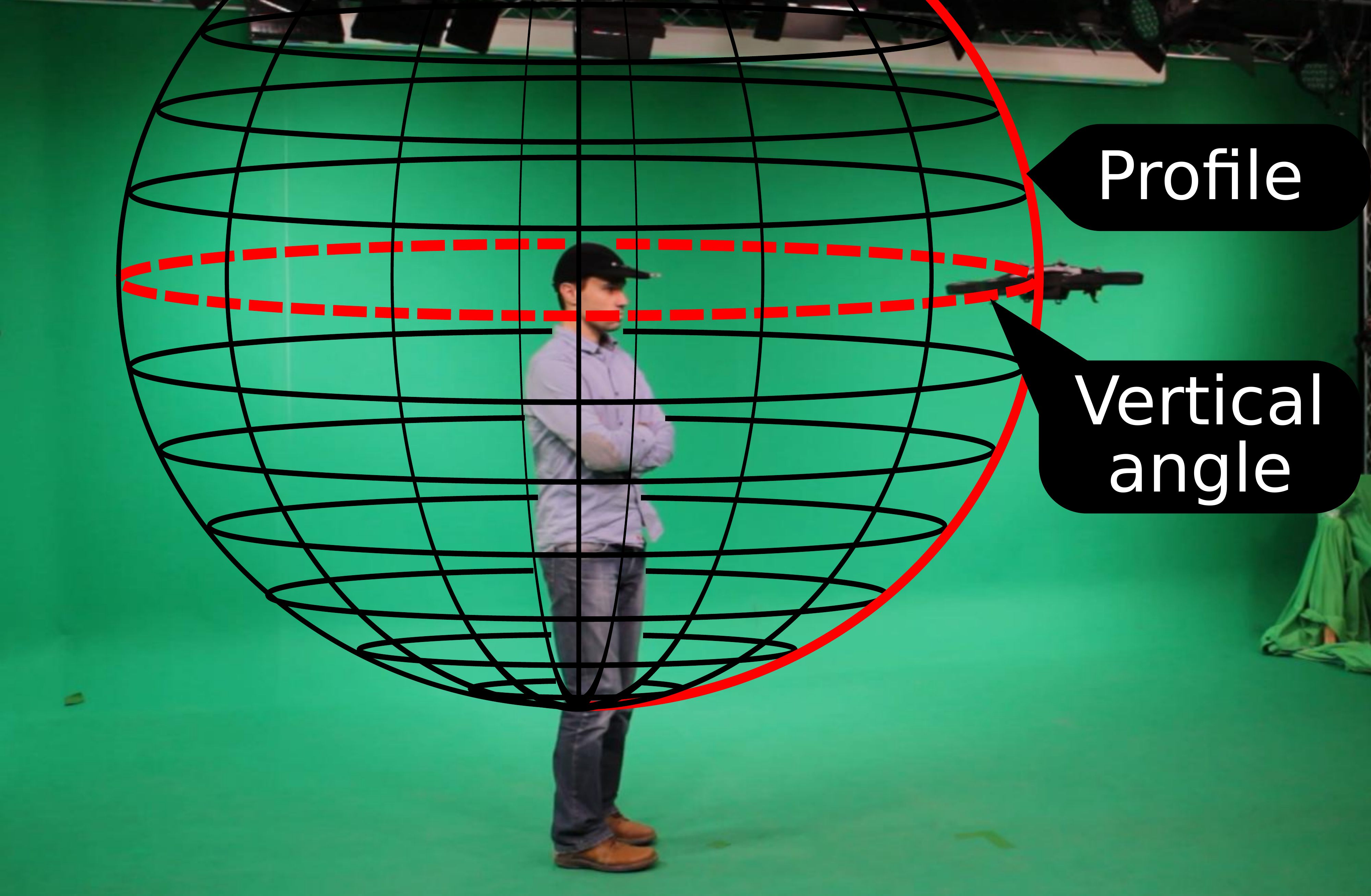}
    	\caption{}
    \end{subfigure}
    \caption{The shot size (a), vertical angle and profile angle (b) define the camera placement for \emph{``MS on A 34left screencenter''}. }
    \label{fig:pslspec}
\end{figure}

Based on the PSL keywords, the pruning process proposed by ~\cite{Galvane2014} gives an interesting solution but does not handle over-constrained PSL specifications.
To solve this issue we propose a different approach.  
Instead of assigning range of possible values on the parametric surface for each of the PSL keywords, we assign exact values and use default ones for unspecified properties.
To solve conflicts in PSL specifications, we ignore the latest conflicting constraint and resume the process.
This rule based approach allows to find exact camera placement for any user input.
Figure~\ref{fig:pslspec} illustrates the placement of a camera for a given PSL specification. The shot size defines the spherical surface while the vertical and profile angle constraints give the position on the surface. The on-screen position only affects the camera orientation.
 
\subsection{Generating the trajectory}
\label{pathGeneration}

\begin{figure}[ht]
    \centering
    \begin{subfigure}[b]{0.49\linewidth}
        \includegraphics[width=\linewidth]{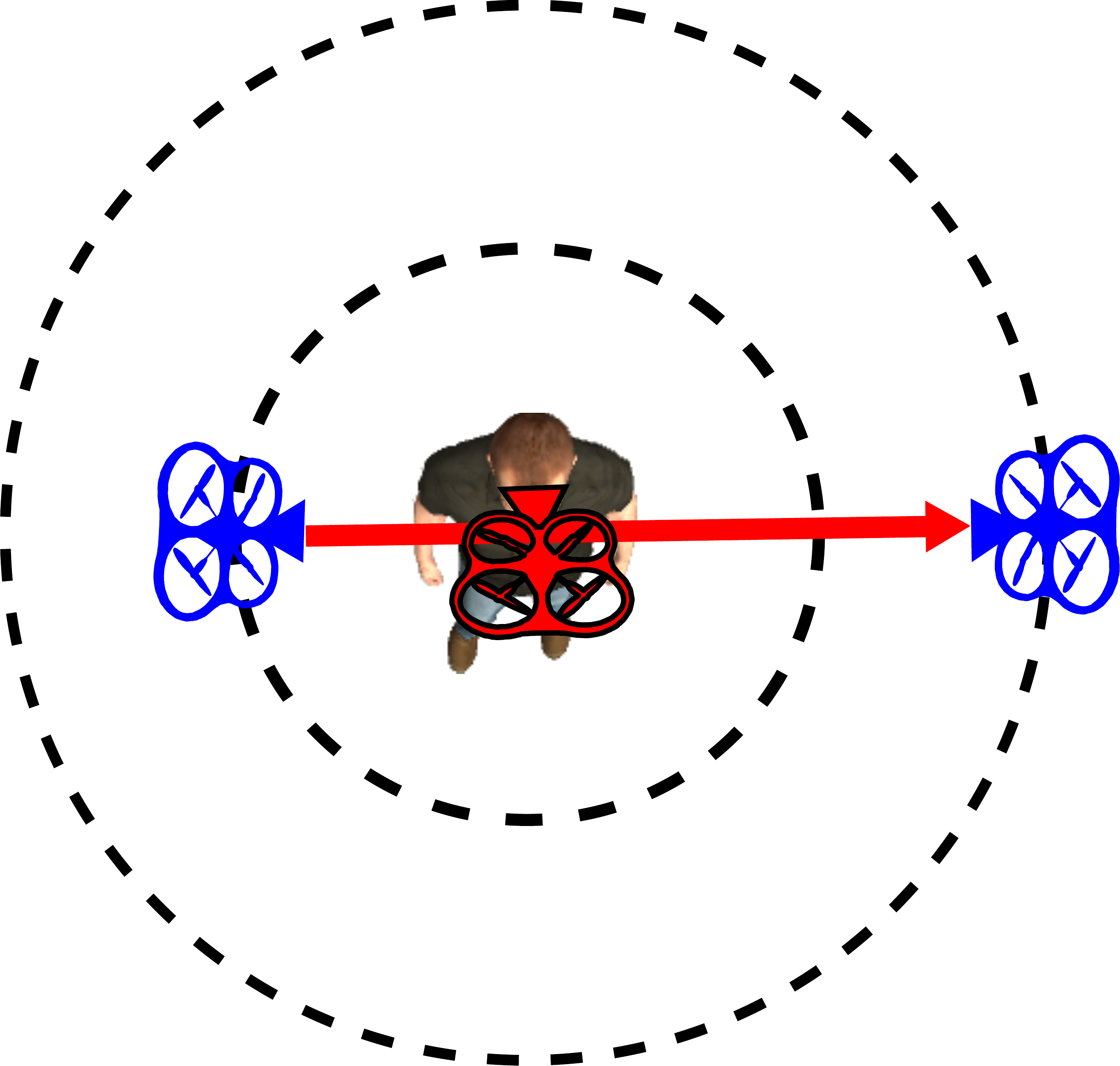}
        \caption{}
        \label{fig:1targetFail}
    \end{subfigure}\hfill
    \begin{subfigure}[b]{0.49\linewidth}
        \includegraphics[width=\linewidth]{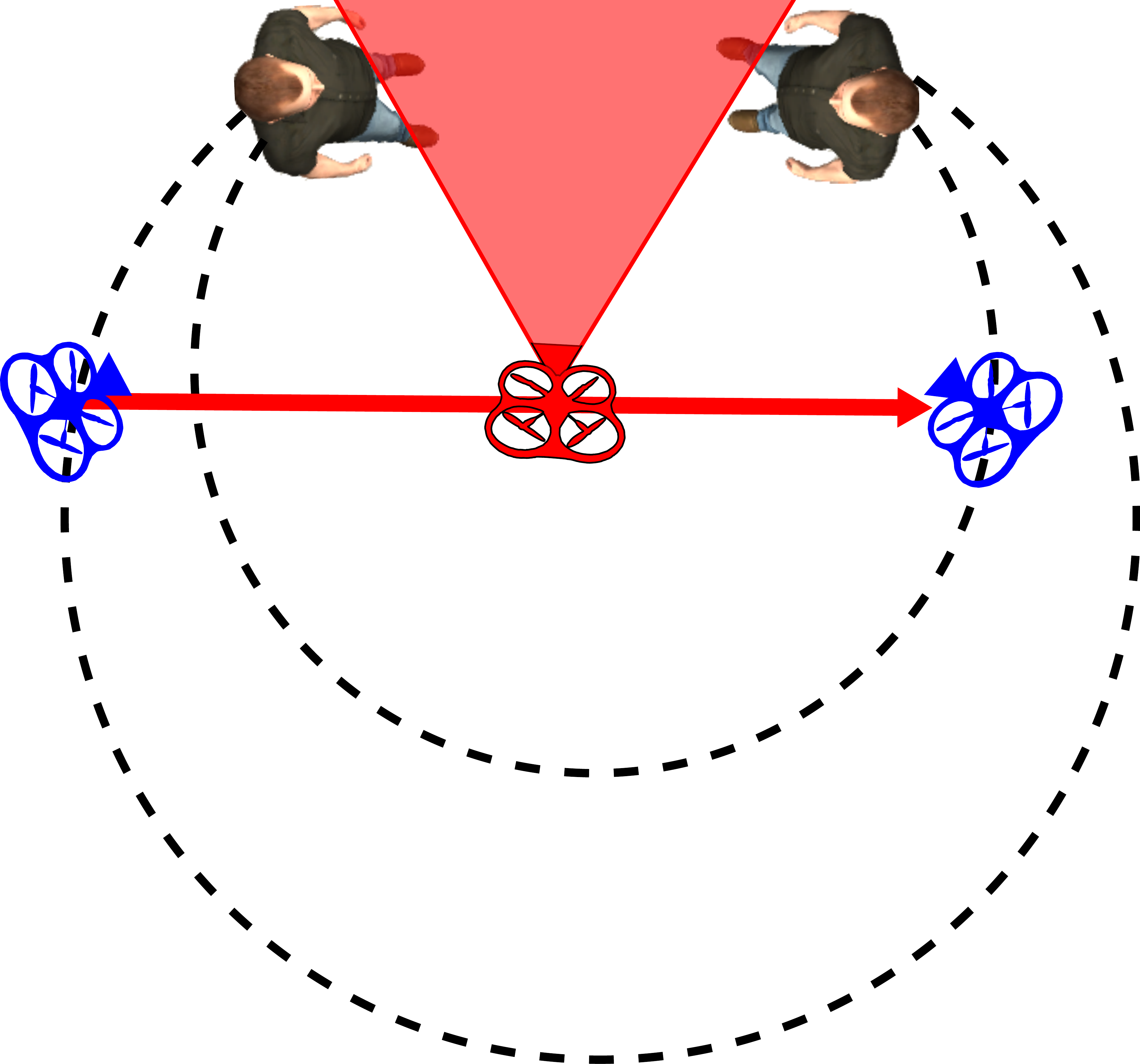}
        \caption{}
        \label{fig:2targetsFail}
    \end{subfigure}
    \caption{Linear interpolations of the 3D world coordinates produces poor trajectories (a) sometimes unable to ensure the framing of the actors (b).}
    \label{fig:interpolation}
\end{figure}

The next step consists of generating a feasible trajectory to move the drone from its current location towards the shot configuration specified by the user.
As shown in Figure~\ref{fig:1targetFail} and~\ref{fig:2targetsFail}, a straightforward linear interpolation of the positions in the 3D world produces poor trajectories where the drone is unable to maintain the framing of the targets.

In order to transition between two camera configurations, \cite{Galvane2015} proposed a solution based on the interpolation of the framing properties.
While dedicated to offline camera path planning -- where the motion of the targets is known in advance --, part of their solution remains pertinent for real-time applications.
To adapt it, the first step consists in computing the initial camera configuration from the current position and orientation of the drone with regards to the designated target. 
Then, given the starting time $t_0$ and final time $t_f$ of the shot -- the user can specify the duration of the shot or the desired average speed of the drone -- we can interpolate each of the visual properties values $P_i$ in the manifold space at time $t$
from the initial camera specification and the user defined specification:

\begin{equation*}
\begin{split}
P_i(t) = P_i(t_0) * \zeta(\dfrac{(t_f - t)}{t_f - t_0}, 0,1) + P_i(t_f) * \zeta(\dfrac{(t - t_0)}{t_f - t_0},0,1) \\
\forall i \in{\{screen~position, vertical~angle, profile~angle, size\}}
\end{split}
\end{equation*}
Where the function $\zeta(x,min,max)$ clamps the value of $x$ between $min$ and $max$.
\begin{figure}[ht]
    \centering
    \begin{subfigure}[b]{0.49\linewidth}
        \includegraphics[width=\linewidth]{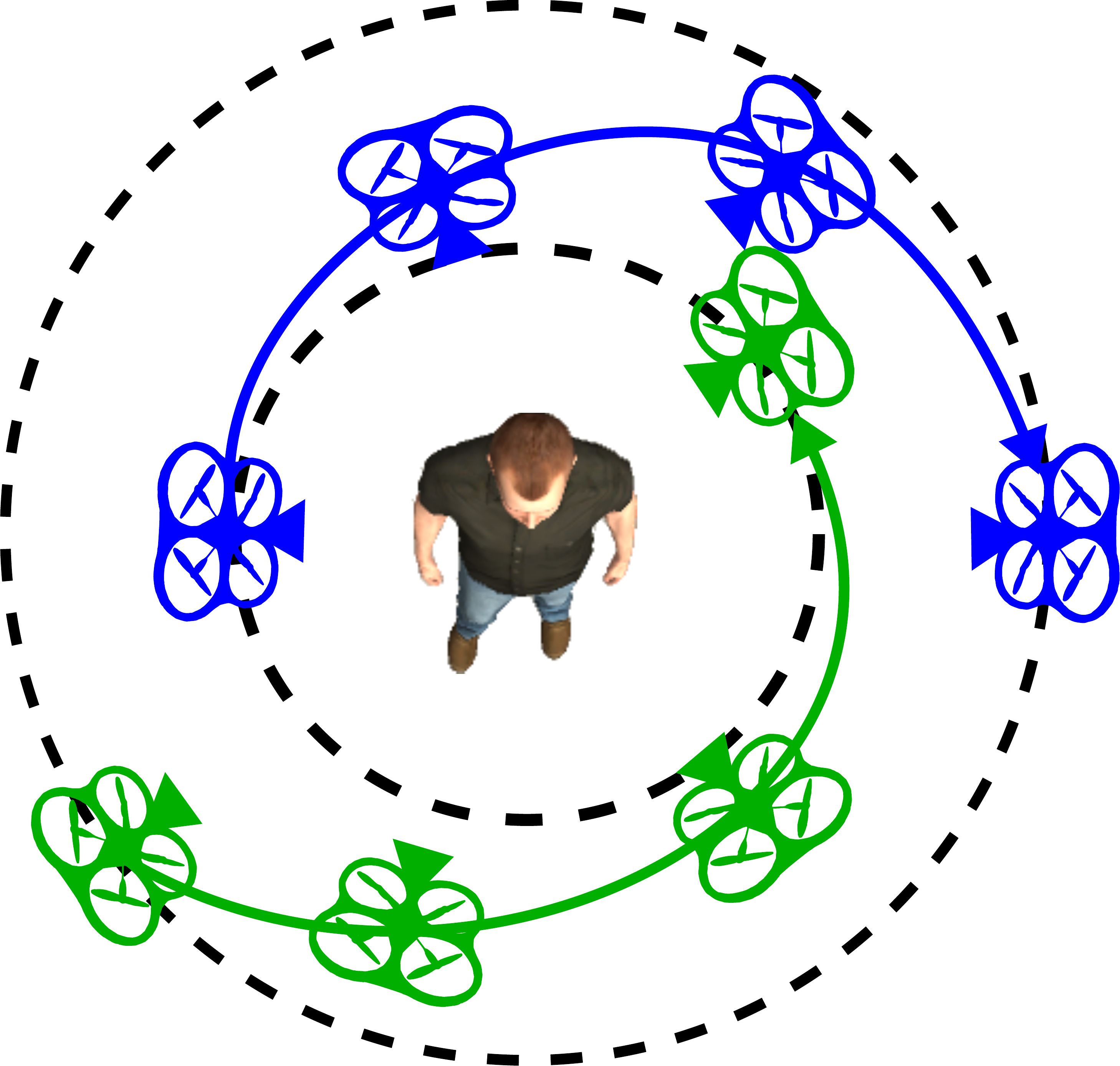}
        \caption{}
        \label{fig:1targetSuccess}
    \end{subfigure}\hfill
    \begin{subfigure}[b]{0.49\linewidth}
        \includegraphics[width=\linewidth]{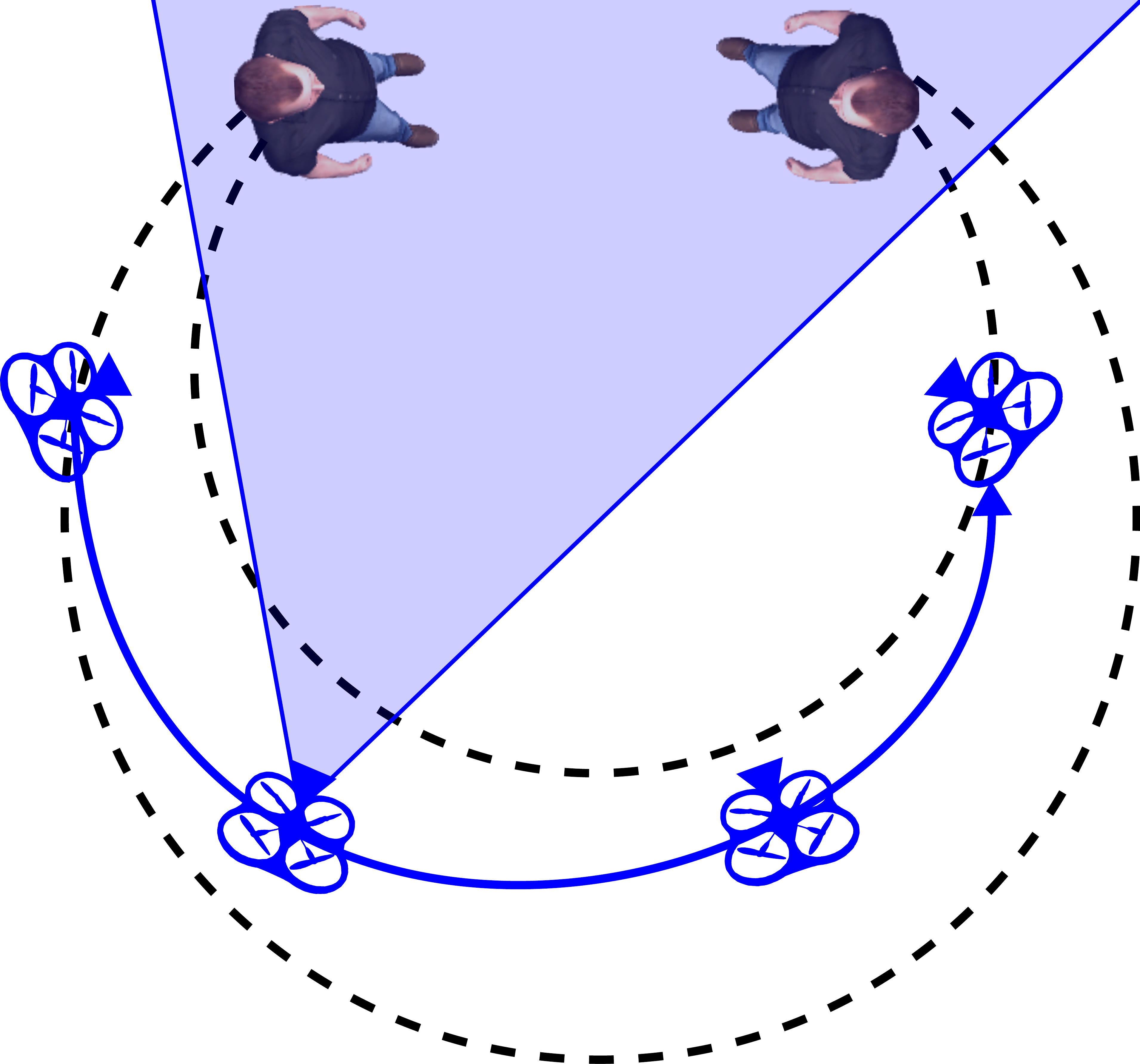}
        \caption{}
        \label{fig:2targetsSuccess}
    \end{subfigure}
    \caption{The interpolation of the framing properties produces natural transitions (a) and allows to maintain visibility properties along the path (b).}
    \label{fig:interpolation2}
\end{figure}

As shown in Figure~\ref{fig:1targetSuccess} and~\ref{fig:2targetsSuccess}, the resulting trajectories provide natural camera motions while ensuring the proper framing of the shot.

\subsection{Navigating along the trajectory}
\label{pathFollowing}

The servo-control loop performed by the \emph{Controller} (detailed in section~\ref{servoControl}) takes as input navigation data that includes the position, speed and course (\emph{i.e.} orientation) of the drone. 
The trajectory computed in the previous stage only allows to access the desired position and orientation of the drone at a given moment in time. It does not provide information on the speed nor manages the acceleration to produce ease in and ease out camera motion.
Drawing inspiration from the steering behaviors introduced in \cite{Galvane2013}, we propose a four stages process that continuously loops to compute a smooth and natural motion along the path until the trajectory is completed. This navigation loop is defined as follows:

\begin{enumerate}
  \item The \emph{Navigator} retrieves and stores the current position data from the \emph{Tracker}. It computes the current velocity from the current and previous positions.
  \item Using the initial camera specification and the user specified configuration, the system computes the targeted position, as described in section~\ref{pathGeneration}.
  \item Based on the drones current position and velocity, we compute the steering forces that will push the drone towards its objective position along the trajectory, while avoiding obstacles (see~\cite{Galvane2014} for implementation details).
  \item The system computes the desired velocity, position and course, and sends this navigation data to the \emph{Controller}.
\end{enumerate}

\section{Drone-independent servo control}
\label{servoControl}
The challenge that we tackle here is the problem of automatic navigation for a generic rotary-wing drone. 
Out-of-the box, such drones usually offer a generic way to manually control their trajectory by adapting four different parameters, namely: the pitch angle $\theta$ (to move forward and backward), the roll angle $\phi$ (to go left and right), the yaw speed $\dot{\psi}$ (to turn around the vertical axis) and the elevation speed $\dot{z}$ (to move up and down).

We here describe a generic control system that uses the current measures of the dynamics (given by the \emph{Tracker})) to adapt the current flight controls (\emph{i.e.} $\theta$, $\phi$, $\dot{\psi}$ and $\dot{z}$) so that the resulting trajectory of the drone follows as closely as possible the one given by the \emph{Navigator}.
The method relies on the two following assumptions:
\begin{enumerate}
\item The roll and pitch angles variations are negligible regarding the yaw angle variations. 
\item The dynamics of the drone in terms of rotation around the up axis are designed to be much slower than the dynamics of the drone in terms of translation along its forward and right axis. 
\end{enumerate}

Under assumption (1), the orientation of the drone in the global frame can be restricted to the only course angle c(t) (linearly related to the yaw angle).
Considering assumption (2), the control of the translation and of the course may be described by two different but coupled linear State Space Representations.
Therefore we propose a compound and coupled model of a generic rotary-wing drone which gives a first-order temporal relation between its flight control and its dynamics. It is composed of:
\begin{itemize}
\item an Explicit Discrete Time-Variant State Space Representation of the translation control of the UAV
\item an Explicit Discrete Time-Variant State Space Representation of the course control of the UAV
\end{itemize}

The global control architecture integrating the two previous models and a Full State
Feedback strategy is based on Kalman filters and account for modeling and measurement errors by using independent and identically distributed additive centered Gaussian noise.
The technical details of our solution are fully disclosed in the patent \cite{Fleureau2015}.

\section{Live platform and training simulator}

\begin{figure}[ht]
	\centering
	\includegraphics[width=\linewidth]{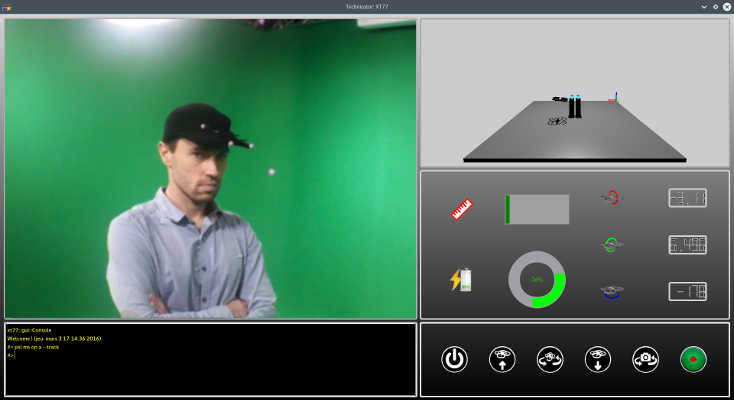}
	\caption{User interface of the framework. It is composed of 5 panels allowing to preview the shot, monitor the state of the drone and interact with it.}
	\label{fig:userInterface}
\end{figure}

Part of this research aimed at providing users with an intuitive tool allowing to easily produce camera shots with no or very little cinematographic and robotic knowledge. To address this challenge, we devised a complete framework that includes a live preview of the shot, a 3D rendering of the scene and two high level control panels.
Figure~\ref{fig:userInterface} shows the resulting user interface.
In addition, we also developped a training platform to simulate the behavior of the drones and allow users to experiment with our tool and its various functionalities before actually shooting a scene.

\begin{figure}[ht]
    \centering
    \begin{subfigure}[b]{0.49\linewidth}
        \includegraphics[width=\linewidth]{images/demo2}
        \caption{}
    \end{subfigure}
    \begin{subfigure}[b]{0.49\linewidth}
        \includegraphics[width=\linewidth]{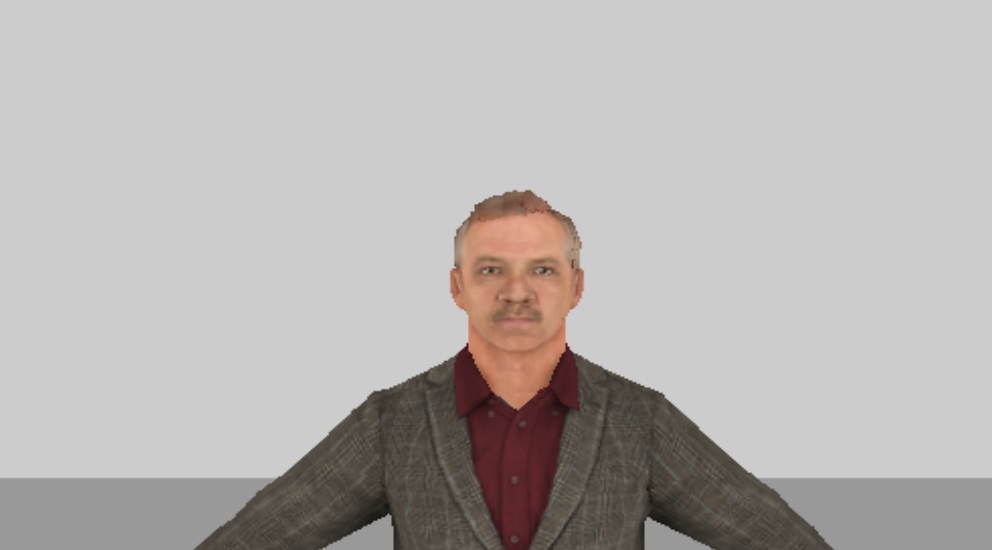}
        \caption{}
    \end{subfigure}
    \caption{Shot preview during the live shooting of a scene (a) and with the rendering of the training simulator (b).}
    \label{fig:preview}
\end{figure}

Divided in five parts, the interface was devised to be as user-friendly as possible.
On the left side, we display the camera output. This allows the user to preview in real-time the shot being produced. 
As shown in Figure~\ref{fig:preview}, in the simulator the video output is replaced by a 3D rendering of the scene generated from the simulated drone.

\begin{figure}[ht]
    \centering
    \begin{subfigure}[b]{0.8\linewidth}
        \includegraphics[width=\linewidth]{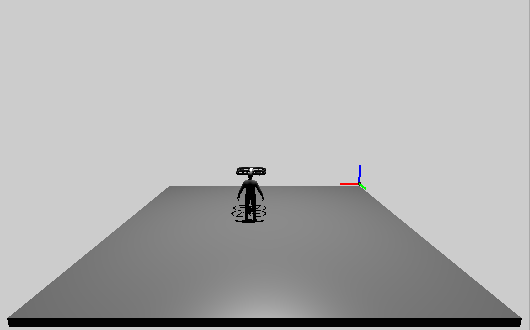}
        \caption{}
        \label{fig:3Dscene}
    \end{subfigure}
    \begin{subfigure}[b]{.8\linewidth}
        \includegraphics[width=\linewidth]{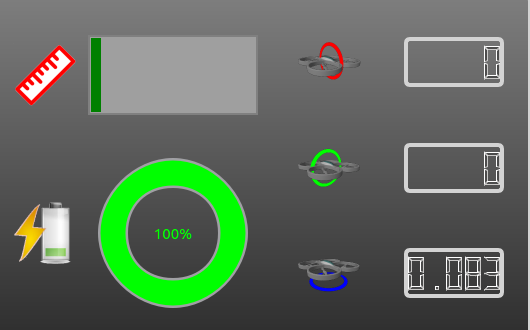}
        \caption{}
        \label{fig:Monitor}
    \end{subfigure}
    \caption{The two information panels display the status of the drone. Position of the drone in the scene (a) ; Status of the drone (b)}
    \label{fig:status}
\end{figure}

On the right side we display all the information relative to the drone. Its position in the 3D world is shown through a 3D rendering of the scene (see Figure~\ref{fig:3Dscene}).
The information shown in Figure~\ref{fig:Monitor} allows to monitor the behaviour of the drone. It shows the error distance to the navigation position, the battery level and the orientation of the drone. 

\begin{figure}[ht]
    \centering
    \includegraphics[width=\linewidth]{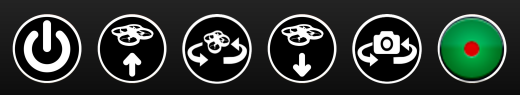}
    \caption{Basic commands: Turn-on/off, take-off, switch drone, land, switch camera and record}
    \label{fig:controls}
\end{figure}

\begin{figure}[ht]
    \centering
    \includegraphics[width=\linewidth]{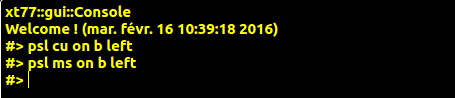}
    \caption{Command interpreter}
	\label{fig:pslInterpreter}
\end{figure}

Finally, the last two components of the interface are the bottom control panels. The right panel shown in Figure~\ref{fig:controls} provides the basic controls of the drones. It manages the primary commands such as turning on and off the drone, taking-off, landing or video recording.
The left panel shown in Figure~\ref{fig:pslInterpreter}  is the main interacting component of this interface. It is the window through which users can communicate their PSL orders. This command interpreter also allows to load predefined trajectories or scripted sequences of commands.

\section{Experimental Results}

In this section we present the results of experiments conducted with a \emph{Parrot AR.Drone}\footnote{http://ardrone2.parrot.com/}.
Our platform was designed to be generic and could be used with any rotary-wing UAV. The \emph{Parrot} was chosen due to its good trade-off between stability, cost and equipment, as well as for the sake of safety. The drone is equipped with a static camera and a safety hull. The tracking of the drones and actors was handled through motion capture with the \emph{OptiTrack} system. 

We tested our platform on a variety of scenarios involving one or two targets, static or moving, and with different user inputs.
All of our results can be seen in the companion video \footnote{https://vimeo.com/157138672}.

\subsection{Single actor scenario}
\begin{figure}[ht]
    \centering
    \begin{subfigure}[b]{0.32\linewidth}
        \includegraphics[width=\linewidth]{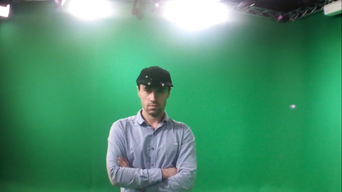}
        \caption{}
    \end{subfigure}
    \begin{subfigure}[b]{.32\linewidth}
        \includegraphics[width=\linewidth]{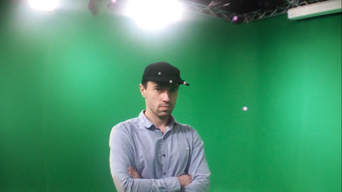}
        \caption{}
    \end{subfigure}
    \begin{subfigure}[b]{.32\linewidth}
        \includegraphics[width=\linewidth]{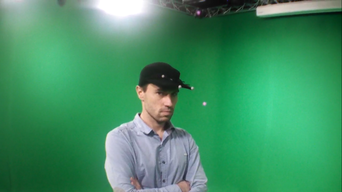}
        \caption{}
    \end{subfigure}

    \begin{subfigure}[b]{0.32\linewidth}
        \includegraphics[width=\linewidth]{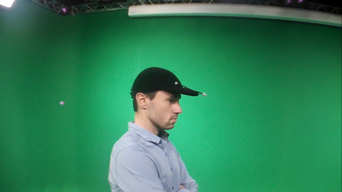}
        \caption{}
    \end{subfigure}
    \begin{subfigure}[b]{.32\linewidth}
        \includegraphics[width=\linewidth]{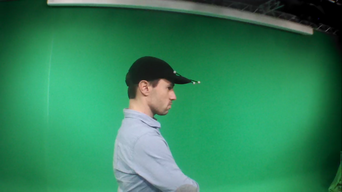}
        \caption{}
    \end{subfigure}
    \begin{subfigure}[b]{.32\linewidth}
        \includegraphics[width=\linewidth]{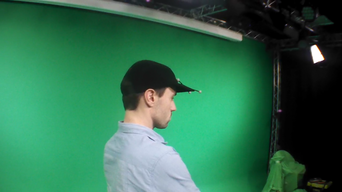}
        \caption{}
    \end{subfigure}
    \caption{The drone transitions from an initial configuration \emph{``MS on A front''} (a) towards a final PSL specification \emph{``MS on A 34backright''} (f)}
    \label{fig:scenario1}
\end{figure}

The first experiment we conducted involved a single actor. Receiving a list of PSL command, the drone executed successively each of the corresponding trajectories and managed to always maintain the framing of the actor.  
Figure~\ref{fig:scenario1} illustrate one of the transition performed by the drone. It shows how the drone is able to smoothly transition from a \emph{front medium shot} to a \emph{3/4 back profile shot}.

\begin{figure}[ht]
    \centering
    \begin{subfigure}[b]{0.32\linewidth}
        \includegraphics[width=\linewidth]{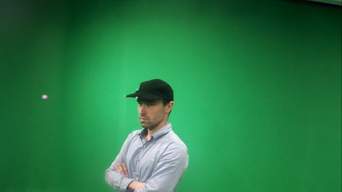}
        \caption{}
    \end{subfigure}
    \begin{subfigure}[b]{.32\linewidth}
        \includegraphics[width=\linewidth]{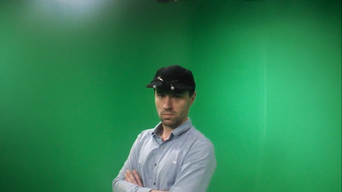}
        \caption{}
    \end{subfigure}
    \begin{subfigure}[b]{.32\linewidth}
        \includegraphics[width=\linewidth]{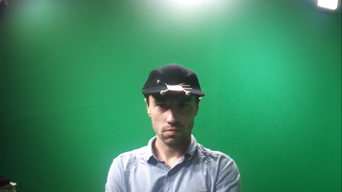}
        \caption{}
    \end{subfigure}

    \begin{subfigure}[b]{0.32\linewidth}
        \includegraphics[width=\linewidth]{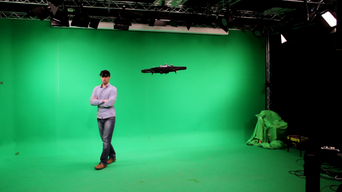}
        \caption{}
    \end{subfigure}
    \begin{subfigure}[b]{.32\linewidth}
        \includegraphics[width=\linewidth]{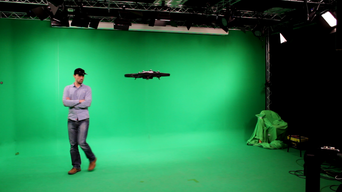}
        \caption{}
    \end{subfigure}
    \begin{subfigure}[b]{.32\linewidth}
        \includegraphics[width=\linewidth]{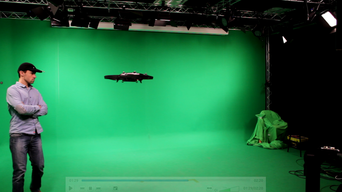}
        \caption{}
    \end{subfigure}
    \caption{The drone autonomously maintains a given framing over a moving target. (a,b,c) show the resulting shot and (d,e,f) give the associated overview of the scene to show the movement of the actor and the drone.}
    \label{fig:scenario2}
\end{figure}

One of the functionalities of our tool is to autonomously maintain the framing of a moving target. 
After validating shot transitions over a static target, we tested the system with a moving actor.
Figure~\ref{fig:scenario2} illustrates the behavior of the drone. It successfully maintains a given framing as the character moves.

\subsection{Two actors scenario}

\begin{figure}[ht]
    \centering
    \begin{subfigure}[b]{0.32\linewidth}
        \includegraphics[width=\linewidth]{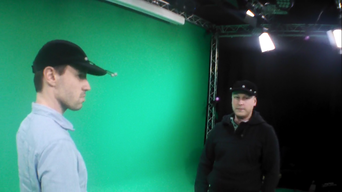}
        \caption{}
    \end{subfigure}
    \begin{subfigure}[b]{.32\linewidth}
        \includegraphics[width=\linewidth]{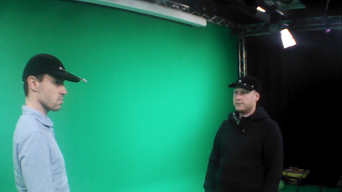}
        \caption{}
    \end{subfigure}
    \begin{subfigure}[b]{.32\linewidth}
        \includegraphics[width=\linewidth]{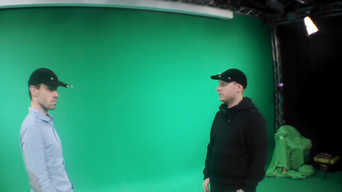}
        \caption{}
    \end{subfigure}

    \begin{subfigure}[b]{0.32\linewidth}
        \includegraphics[width=\linewidth]{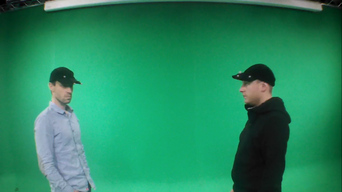}
        \caption{}
    \end{subfigure}
    \begin{subfigure}[b]{.32\linewidth}
        \includegraphics[width=\linewidth]{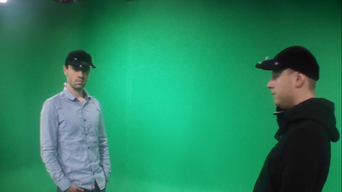}
        \caption{}
    \end{subfigure}
    \begin{subfigure}[b]{.32\linewidth}
        \includegraphics[width=\linewidth]{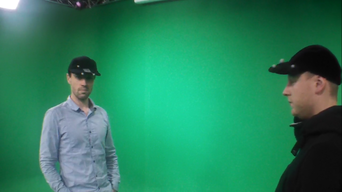}
        \caption{}
    \end{subfigure}
    \caption{The drone transitions from an initial configuration \emph{``MS on A screenleft and B screenright''} (a) towards a final PSL specification  \emph{``MS on B screenright and A screenleft''}(f)}
    \label{fig:scenario3}
\end{figure}

After testing our framework on a single-character scenario, we extended the experiment by adding another actor.
Figure~\ref{fig:scenario3} shows a transition performed from an \emph{Over-the-shoulder} shot to its opposite \emph{Over-the-shoulder} shot.
	
\begin{figure}[ht]
    \centering
    \begin{subfigure}[b]{0.32\linewidth}
        \includegraphics[width=\linewidth]{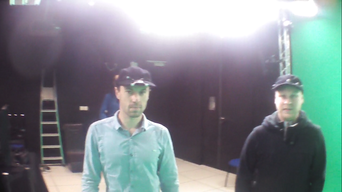}
        \caption{}
    \end{subfigure}
    \begin{subfigure}[b]{.32\linewidth}
        \includegraphics[width=\linewidth]{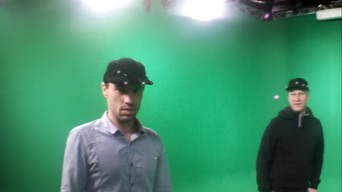}
        \caption{}
    \end{subfigure}
    \begin{subfigure}[b]{.32\linewidth}
        \includegraphics[width=\linewidth]{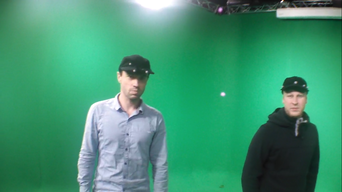}
        \caption{}
    \end{subfigure}

    \begin{subfigure}[b]{0.32\linewidth}
        \includegraphics[width=\linewidth]{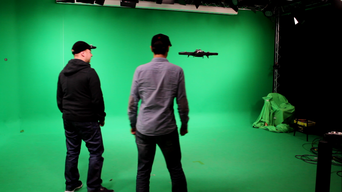}
        \caption{}
    \end{subfigure}
    \begin{subfigure}[b]{.32\linewidth}
        \includegraphics[width=\linewidth]{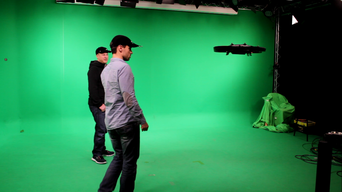}
        \caption{}
    \end{subfigure}
    \begin{subfigure}[b]{.32\linewidth}
        \includegraphics[width=\linewidth]{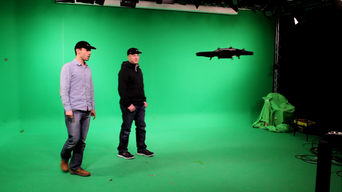}
        \caption{}
    \end{subfigure}
    \caption{The drone autonomously maintains a given framing over two moving actors. (a,b,c) show the resulting shot and (d,e,f) give the associated overview of the scene to show the movement of the actors and the drone.}
    \label{fig:scenario4}
\end{figure}

Once again, after testing the system with static targets, we asked the actors to move around in the scene.
Figure~\ref{fig:scenario4} illustrates part of this experiment. It shows that the system continuously manages to maintain the screen composition of the shot even though the two characters move at a different speed and in different directions.

\section{Limitations and Future work}

One of the main limitations of the tool is the lack of efficient obstacle avoidance mechanism. 
For safety reason, such feature will have to be implemented before testing the system with heavier and unprotected drones.
The work by~\cite{Oskam2009} constitutes an interesting lead. It introduced an effective solution for the computation of collision-free camera trajectories in real-time.

Controlling several drones in real-time represents another interesting challenge. Even though the task of managing multiple drones has already been addressed on several occasions \cite{Kushleyev2013, Mellinger2010}, it never dealt with issues relative to camera shooting such as the visibility.
Moreover, handling several drones simultaneously opens the way to new possibilities. For instance, the investigation of live-editing solutions \cite{Lino2011, Christie2012, He1996} might offer interesting options to improve our framework.

Finally, in order to further validate our approach, we will be conducting user studies involving cinematography experts, drone pilots and inexperienced user.

\section{Conclusions}

 \begin{quotation}
\emph{One of the ultimate goals in Robotics is to create autonomous robots. Such robots will accept high-level descriptions of tasks and will execute them without further human intervention. The input descriptions will specify what the users wants rather than how to do it. The robots will be any kind of versatile mechanical device equipped with actuators and sensors under the control of a computing system.}

\hfill --- Robot Motion Planning, \emph{J.C. Latombe}. 
\end{quotation}

Throughout this research, we have tackled this challenge of creating autonomous robots. We addressed the task, traditionally carried out by a cameraman, of producing live cinematographic shots given a high-level description of the desired result (usually specified by a director).

Adapting virtual camera control techniques to handle the task of navigating an UAV in a constrained environment, we devised an intuitive tool that autonomously handles the difficult task of maneuvering a drone in real-time to maintain a precise frame composition over moving targets.
Furthermore, the solution that we propose is drone-independent and can be used with any rotary-wing UAV offering generic flight control.

Finally, the first series of experiments conducted with our framework gave promising results that will be further validated through several user studies.



\bibliographystyle{eg-alpha-doi}

\bibliography{wiced2016}

\newcommand{\etalchar}[1]{$^{#1}$}
\begin{thebibliography}{\uppercase{GBFGCL12}}

\bibitem[3Dr]{3Drsolo}
3drsolo.
\newblock \url{https://3dr.com/}.

\bibitem[Ari76]{Arijon1976}
\textsc{Arijon D.}:
\newblock \emph{Grammar of the Film Language}.
\newblock Silman-James Press, 1976.

\bibitem[CLR12]{Christie2012}
\textsc{Christie M., Lino C., Ronfard R.}:
\newblock Film editing for third person games and machinima.
\newblock In \emph{Workshop on Intelligent Cinematography and Editing} (2012),
  ACM.

\bibitem[CON08]{Christie2008}
\textsc{Christie M., Olivier P., Normand J.-M.}:
\newblock Camera control in computer graphics.
\newblock \emph{Computer Graphics Forum 27}, 8 (2008), 2197--2218.

\bibitem[DJI]{DJI}
Dji.
\newblock \url{www.dji.com}.

\bibitem[FTKLC15]{Fleureau2015}
\textsc{Fleureau J., Tariolle F.-L., Kerbiriou P., Le~Clerc F.}:
\newblock Method for controlling a path of a rotary-wing drone, a corresponding
  system, a rotary-wing drone implementing this system and the related uses of
  such a drone, 12 2015.
\newblock URL: \url{https://www.lens.org/lens/patent/US_2015_0370258_A1}.

\bibitem[GBFGCL12]{Gomez2012}
\textsc{Gomez-Balderas J.~E., Flores G., Garc{\'i}a~Carrillo L.~R., Lozano R.}:
\newblock Tracking a ground moving target with a quadrotor using switching
  control.
\newblock \emph{Journal of Intelligent \& Robotic Systems 70}, 1 (2012),
  65--78.

\bibitem[GCLR15]{Galvane2015}
\textsc{Galvane Q., Christie M., Lino C., Ronfard R.}:
\newblock {Camera-on-rails: Automated Computation of Constrained Camera Paths}.
\newblock In \emph{{ ACM SIGGRAPH Conference on Motion in Games}} (Paris,
  France, Nov. 2015).

\bibitem[GCR{\etalchar{*}}13]{Galvane2013}
\textsc{Galvane Q., Christie M., Ronfard R., Lim C.-K., Cani M.-P.}:
\newblock {Steering Behaviors for Autonomous Cameras}.
\newblock In \emph{{MIG 2013 - ACM SIGGRAPH conference on Motion in Games}}
  (Dublin, Ireland, Nov. 2013), MIG '13 Proceedings of Motion on Games, {ACM},
  pp.~93--102.

\bibitem[GRCS14]{Galvane2014}
\textsc{Galvane Q., Ronfard R., Christie M., Szilas N.}:
\newblock {Narrative-Driven Camera Control for Cinematic Replay of Computer
  Games}.
\newblock In \emph{{Motion In Games}} (Los Angeles, United States, Nov. 2014).

\bibitem[HCS96]{He1996}
\textsc{He L.-w., Cohen M.~F., Salesin D.~H.}:
\newblock The virtual cinematographer: A paradigm for automatic real-time
  camera control and directing.
\newblock In \emph{Proceedings of the 23rd Annual Conference on Computer
  Graphics and Interactive Techniques} (New York, NY, USA, 1996), SIGGRAPH '96,
  ACM, pp.~217--224.

\bibitem[Hex]{Hexo}
Hexo+.
\newblock \url{https://hexoplus.com/}.

\bibitem[JRT{\etalchar{*}}15]{Joubert2015}
\textsc{Joubert N., Roberts M., Truong A., Berthouzoz F., Hanrahan P.}:
\newblock An interactive tool for designing quadrotor camera shots.
\newblock \emph{ACM Trans. Graph. 34}, 6 (Oct. 2015), 238:1--238:11.

\bibitem[Kat09]{Kat1991}
\textsc{Katz S.~D.}:
\newblock \emph{Film Directing Shot by Shot: Visualizing from Concept to
  Screen}.
\newblock Focal Press, 2009.

\bibitem[KMPK13]{Kushleyev2013}
\textsc{Kushleyev A., Mellinger D., Powers C., Kumar V.}:
\newblock Towards a swarm of agile micro quadrotors.
\newblock \emph{Auton. Robots 35}, 4 (Nov. 2013), 287--300.

\bibitem[LC12]{Lino2012}
\textsc{Lino C., Christie M.}:
\newblock Efficient composition for virtual camera control.
\newblock In \emph{Proceedings of the ACM SIGGRAPH/Eurographics Symposium on
  Computer Animation} (2012), Eurographics Association, pp.~65--70.

\bibitem[LC15]{Lino2015}
\textsc{Lino C., Christie M.}:
\newblock Intuitive and efficient camera control with the toric space.
\newblock \emph{ACM Transactions on Graphics (TOG) 34}, 4 (2015), 82.

\bibitem[LCCR11]{Lino2011}
\textsc{Lino C., Chollet M., Christie M., Ronfard R.}:
\newblock Computational model of film editing for interactive storytelling.
\newblock In \emph{Interactive Storytelling}. Springer Berlin Heidelberg, 2011,
  pp.~305--308.

\bibitem[Mas65]{Mas1965}
\textsc{Mascelli J.~V.}:
\newblock \emph{The Five C's of Cinematography: Motion Picture Filming
  Techniques}.
\newblock Silman-James Press, 1965.

\bibitem[Mer10]{Merc2010}
\textsc{Mercado G.}:
\newblock \emph{The Filmmaker's Eye: Learning (and Breaking) the Rules of
  Cinematic Composition}.
\newblock Focal Press, 2010.

\bibitem[Mis]{MissionPlanner}
Mission planner.
\newblock \url{http://planner.ardupilot.com/}.

\bibitem[MK11]{Mellinger2011}
\textsc{Mellinger D., Kumar V.}:
\newblock Minimum snap trajectory generation and control for quadrotors.
\newblock In \emph{Robotics and Automation (ICRA), 2011 IEEE International
  Conference on} (2011), IEEE, pp.~2520--2525.

\bibitem[MMK14]{Mellinger2014}
\textsc{Mellinger D., Michael N., Kumar V.}:
\newblock Trajectory generation and control for precise aggressive maneuvers
  with quadrotors.
\newblock In \emph{Experimental Robotics}. Springer Berlin Heidelberg, 2014,
  pp.~361--373.

\bibitem[MSMK10]{Mellinger2010}
\textsc{Mellinger D., Shomin M., Michael N., Kumar V.}:
\newblock Cooperative grasping and transport using multiple quadrotors.
\newblock In \emph{DARS} (2010), Martinoli A., Mondada F., Correll N., Mermoud
  G., Egerstedt M., Hsieh M.~A., Parker L.~E., Støy K., (Eds.), vol.~83 of
  \emph{Springer Tracts in Advanced Robotics}, Springer, pp.~545--558.

\bibitem[MTFP11]{Meier2011}
\textsc{Meier L., Tanskanen P., Fraundorfer F., Pollefeys M.}:
\newblock Pixhawk: A system for autonomous flight using onboard computer
  vision.
\newblock In \emph{Robotics and Automation (ICRA), 2011 IEEE International
  Conference on} (May 2011), pp.~2992--2997.

\bibitem[MTH{\etalchar{*}}12]{Meier2012}
\textsc{Meier L., Tanskanen P., Heng L., Lee G.~H., Fraundorfer F., Pollefeys
  M.}:
\newblock Pixhawk: A micro aerial vehicle design for autonomous flight using
  onboard computer vision.
\newblock \emph{Autonomous Robots 33}, 1-2 (2012), 21--39.

\bibitem[Mur86]{Murch1986}
\textsc{Murch W.}:
\newblock \emph{In the blink of an eye}.
\newblock Silman-James Press, 1986.

\bibitem[OSTG09]{Oskam2009}
\textsc{Oskam T., Sumner R.~W., Thuerey N., Gross M.}:
\newblock Visibility transition planning for dynamic camera control.
\newblock In \emph{Proceedings of the 2009 ACM SIGGRAPH/Eurographics Symposium
  on Computer Animation} (New York, NY, USA, 2009), SCA '09, ACM, pp.~55--65.

\bibitem[RBR13]{Richter2013}
\textsc{Richter C., Bry A., Roy N.}:
\newblock Polynomial trajectory planning for aggressive quadrotor flight in
  dense indoor environments.
\newblock \emph{Proceedings of the International Symposium on Robotics Research
  (ISRR)} (2013).

\bibitem[RGB13]{Ronfard2013}
\textsc{Ronfard R., Gandhi V., Boiron L.}:
\newblock {The Prose Storyboard Language: A Tool for Annotating and Directing
  Movies}.
\newblock In \emph{{2nd Workshop on Intelligent Cinematography and Editing part
  of Foundations of Digital Games - FDG 2013}} (Chania, Crete, Greece, May
  2013), {Society for the Advancement of the Science of Digital Games}.

\bibitem[SBD14]{Srikanth2014}
\textsc{Srikanth M., Bala K., Durand F.}:
\newblock Computational rim illumination with aerial robots.
\newblock In \emph{Proceedings of the Workshop on Computational Aesthetics}
  (New York, NY, USA, 2014), CAe '14, ACM, pp.~57--66.

\bibitem[TB93]{Thompson1993}
\textsc{Thompson R., Bowen C.~J.}:
\newblock \emph{Grammar of the Edit}.
\newblock Focal Press, 1993.

\bibitem[TB98]{Thompson1998}
\textsc{Thompson R., Bowen C.~J.}:
\newblock \emph{Grammar of the Shot}.
\newblock Focal Press, 1998.

\bibitem[TEM11]{Teuliere2011}
\textsc{Teuliere C., Eck L., Marchand E.}:
\newblock {Chasing a moving target from a flying UAV}.
\newblock In \emph{{IEEE/RSJ Int. Conf. on Intelligent Robots and Systems,
  IROS'11}} (San Francisco, USA, United States, 2011).

\end{thebibliography}

\end{document}